\documentclass{acm_proc_article-sp}
\pdfoutput=1
\usepackage{booktabs}
\usepackage{caption}
\usepackage{censor}
\usepackage{comment}
\usepackage[utf8]{inputenc}
\usepackage{times}
\usepackage{url}
\usepackage{latexsym}
\usepackage{numprint}
\usepackage{hyperref}
\usepackage{color}

\hypersetup{colorlinks=true,citecolor=[rgb]{0,0.6,0}}
\definecolor{whitegray}{rgb}{0.985,0.985,0.975}

\newcommand{\figurename}{Figure}

\title{Wembedder:
  Wikidata entity embedding web service}
\author{
Finn Årup Nielsen \\
  Cognitive Systems, DTU Compute \\
  Technical University of Denmark \\
  Kongens Lyngby, Denmark \\
  faan@dtu.dk \\
}

\begin{document}

\maketitle

\begin{abstract}
  I present a web service for querying an embedding of entities in the
  Wikidata knowledge graph. 
  The embedding is trained on the Wikidata dump using Gensim's Word2Vec
  implementation and a simple graph walk. A REST API is implemented.
  Together with the Wikidata API the web service exposes a
  multilingual resource for over 600'000 Wikidata items and properties.
\end{abstract}


\sloppypar

\section{Introduction}

The Word2Vec model \cite{Q24699014} spawned an interest in dense
word representation in a low-dimensional space, 
and there are now a considerable number of ``2vec'' models beyond the
word level.\footnote{\url{https://github.com/MaxwellRebo/awesome-2vec}}
One recent avenue of research in the ``2vec'' domain uses knowledge
graphs \cite{Q30246651}. 
Such systems can take advantage of the large knowledge graphs, e.g.,
DBpedia or Freebase, for graph embedding. 
Graph embedding in the simplest case would map individual nodes of the
network to a continuous low-dimensional space, while embedding with
knowledge graphs would typically handle the typed links between
knowledge items/nodes.

Wikidata \url{https://www.wikidata.org/} \cite{Q18507561} is a
relatively new knowledge graph resource.
It is run by the Wikimedia Foundation that is also behind Wikipedia, 
thus Wikidata can be regarded as a sister site to Wikipedia.
While Wikipedia has been extensively used as a data and text mining
resource \cite{Q27615040}, 
Wikidata has so far seen less use in machine learning contexts.
There are several advantages with Wikidata.
Wikidata is not tied to a single language, but can include labels for 
hundreds of languages for each item in the knowledge graph.
As such, an embedding that works from Wikidata items is in principle
multilingual (however, there is no guarantee that the item label for a
specific language is set).
Another advantage with Wikidata is that each item can provide extra
contextual data from the Wikidata statements. 
Search in Wikidata is enabled by string-based search engines in
the Wikidata API as well as the SPARQL-based Wikidata
Query Service (WDQS).
General functions for finding related items or generating
fixed-length features for machine learning are to my knowledge not
available.

There is some research that combines machine learning and Wikidata 
\cite{Q27036495,Q31769808}, e.g.,
Mousselly-Sergieh and Gurevych have presented a method for aligning
Wikidata items with FrameNet based on Wikidata labels and aliases
\cite{Q27036495}. 

Property suggestion is running live in the Wikidata editing interface, 
where it helps editors recall appropriate properties for items during
manual editing, 
and as such a form of recommender system.
Researchers have investigated various methods for this process
\cite{Q27044266}. 

Scholia at \url{https://tools.wmflabs.org/scholia/} is our web service
that presents scholarly profiles based on data in Wikidata extracted with
WDQS \cite{Q28942417}.
Counting co-occurrence patterns with SPARQL queries, Scholia can list
related items based on a query item. 
For instance, Scholia lists related diseases based on overlapping
associated genes.\footnote{See, e.g., the page for schizophrenia at
\url{https://tools.wmflabs.org/scholia/disease/Q41112}.} 
Other than these count- and SPARQL-based methods Scholia has limited
means to show related items to a Scholia user.

Several research groups provide word embedding web services: 
GPL-licensed \emph{WebVectors} uses a Flask and Gensim
\cite{Q32132685,Q32138153}, 
and instances for English and Russian run at 
\url{http://rusvectores.org/} and for English and
Norwegian at \url{http://vectors.nlpl.eu/explore/embeddings/}.
A Turku BioNLP group provides a Flask-based word embedding web
service at
\url{http://bionlp-www.utu.fi/wv_demo/}
based on the English Google News and Finnish Corpora.
A web service for handling multilingual word embeddings has also been
announced \cite{Q32129681}. 
Wembedder is distinguished from these services by using the Wikidata
entities (items and properties) as the ``words'' in the embedding (rather than
natural language words) and by using
the live Wikidata web service to provide multilingual labels for the
entities.

\section{Wembedder}

\begin{figure}[tb]
  \centering
  \includegraphics[width=\columnwidth]{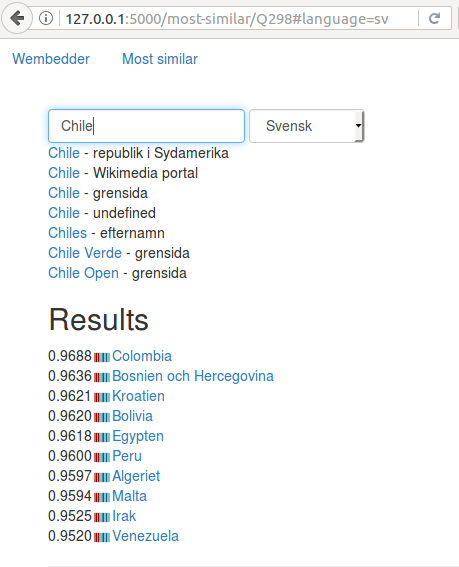}
  \caption{Wembedder's output after a query on Q298 (the country
    Chile) with the interface set to Swedish.  
  }
  \label{fig:chile}
\end{figure}

\subsection{Model setup}

The Wikimedia Foundation provides the Wikidata RDF dumps for download
at \url{https://dumps.wikimedia.org/wikidatawiki/entities/}.
For the setup of the initial model,
I downloaded the so-called truthy
dumps available in Notation3 format. 
The specific file was the 5.2~GB large compressed file
\texttt{wikidata-20170613-truthy-BETA.nt.bz2}.  
The truthy dumps only have a limited set of all the triples in
Wikidata: 
Those that are associated with the \verb!wdt! prefix.
From this dump, I extracted the triples where both subject and object
were Wikidata items, i.e., leaving out triples where the object is a
value such as an identifier, a date, a name, etc.
The generated file contains 88'941'173 lines each with a triple.
The \texttt{http://www.wikidata.org/entity/} and
\texttt{http://www.wikidata.org/prop/direct/} prefixes were stripped, 
so the first few lines of the generated file have the following
content in a format similar to Magnus Manske's QuickStatements format:
\begin{verbatim}
Q22 P1546 Q2016568
Q22 P610 Q104674
Q22 P1151 Q8143311
Q22 P31 Q3336843
Q22 P36 Q23436
Q22 P47 Q21
...
\end{verbatim}
Each line can be regarded as a very simple graph walk consisting of a
single step from one Wikidata item through a typed property to the
next Wikidata item.
These triple data I now regard as a sentence of three ``words'' which
can be treated by standard word embedding implementations.
I use the Word2Vec model in the Gensim program \cite{Q28042398}. 
The initial model trained used the CBOW training algorithm, an embedding
dimension on 100, a 
window of 1 and a minimum count of 20, i.e., any ``word'' must appear
20 times or more to be included in the model.
The rest of the parameters in the Word2Vec model were kept at the
Gensim defaults.
With this setup, the model ends up with a vocabulary of 609'471.
This number includes 738 properties and 608'733 Wikidata items.
Gensim can store its model parameters in files with a combined size of
518 megabytes.
A permanent version of the model parameters is available in Zenodo
under DOI
\href{https://dx.doi.org/10.5281/zenodo.823195}{10.5281/zenodo.823195}.

% len([key for key in w2v.wv.vocab.keys() if key.startswith('P')])

\subsection{Web service}

The web service was set up with the Python Flask web framework
\cite{Q28822647} with the Apache-licensed code available at a GitHub
repository: \url{https://github.com/fnielsen/wembedder}.
\figurename~\ref{fig:chile} shows the interface.
A version of Wembedder runs from
\url{https://tools.wmflabs.org/wembedder/}, i.e., from the cloud
service provided by the Wikimedia Foundation.

The Wembedder web service relies on the Wikidata API at
\url{https://www.wikidata.org/w/api.php}
and its \emph{wbsearchentities} action for searching for
items in multiple languages in an implementation based on the search
facility on the Wikidata homepage.
Labels for searching and labels for the results are generated via ajax
calls to the Wikidata API. 
\begin{comment}
This may be slightly confusing to the user as the search interface
shows items that are out-of-vocabulary, and when these are queried
Wembedder returns an empty result.
\end{comment}

\begin{figure}[tb]
  \centering
  \includegraphics[width=\columnwidth]{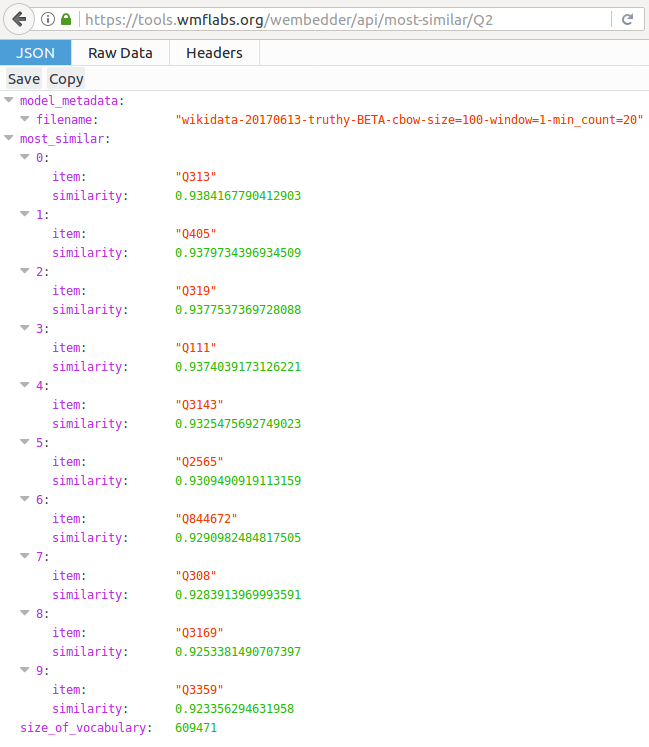}
  \caption{JSON output from Wembedder's REST API with a query on Q2
    (Earth) rendered in a web browser.
    The first entities in the result list is Q313 (Venus).
  }
  \label{fig:api}
\end{figure}

A REST API is implemented as part of Wembedder and returns
JSON-formatted results, e.g., 
\verb!/api/most-similar/Q80! will return the most
similar entities for a query on Tim Berners-Lee (Q80), 
see also \figurename~\ref{fig:api}.
Similarity computations are implemented with a URL such as
\verb!/api/similarity/Q2/Q313!. 
Here the Earth and Venus are compared.  
The human interface of Wembedder uses the REST API in a ajax fashion, 
returning an HTML page with an empty result list and with JavaScript
for the actual fetching of the results.

\section{Evaluation}

The embedding in the current version of Wembedder is fairly 
simple compared to the state of the art embeddings, 
that uses complex/holo\-graphic knowledge graph embedding \cite{Q32000242} 
or multiple knowledge graphs and pre-trained corpora-based resources
for building the embedding \cite{Q31895639}.
One should not expect Wembedder to perform at the state of the art
level,\footnote{An overview of
  the state of the art performance in semantic relatedness task,
  including for the Wordsim-353 task, is available at
  \url{https://aclweb.org/aclwiki/index.php?title=Similarity_(State_of_the_art)}}   
and a comparison with the Wordsim-353 dataset for semantic relatedness
evaluation \cite{Q28045598} shows poor performance with Pearson and Spearman
correlations on just 0.13. 

When used to evaluate the Wikidata graph embedding, a matching is
needed between English Wordsim-353 words and the Wikidata items.
It is not straightforward as there usually is
a semantic difference between the words and the items.
It is often possible to find the word as the English label in a Wikidata
item, but for instance, for the Wordsim-353 word
``Japanese'' one must decide whether it should be linked to Japanese
as a language (\href{https://www.wikidata.org/wiki/Q5287}{Q5287}), 
Japanese as a people (\href{https://www.wikidata.org/wiki/Q161652}{Q161652}),
another item (e.g., the disambiguation page,
\href{https://www.wikidata.org/wiki/Q346080}{Q346080}) or an average
or sum over the items.
I attempted to match the words with items, but left several unmatched so
only 278 of the word pairs of the 353 were possible in the analysis.
The correlations were computed from these 278 word pairs.
A skipgram trained model yielded even lower performance with
correlations of just 0.11 and 0.10 for Pearson and Spearman
correlations, respectively.
A CBOW model trained with a higher number of iterations (DOI
\href{https://doi.org/10.5281/zenodo.827339}{10.5281/zenodo.827339})
performed somewhat better with correlations of 0.21.

\section{Discussion and future work}

Wembedder---with its 100-dimensional Gensim model query---will usually be
able to return results in around one second, while the API call is
considerably faster.
It means that it could be used for interactive ``related items''
search.
The SPARQL-based related items queries in Scholia usually takes 
several seconds.

Wikidata at its current state is mostly an encyclopedic source
having little lexical information. 
State of the art relational modeling 
ConceptNet is setup from both encyclopedic
and lexical knowledge graphs as well as corpus-based embeddings
\cite{Q31895639}.
Embeddings based on Wikidata could presumably perform better by using
the link to Wikipedia with the different language versions of
Wikipedia acting as a corpora.
There exist several works describing joint models of words and
entities from knowledge bases/graphs, see, e.g., \cite{Q30095029} and
reference therein.
There is work underway to enable Wikidata to represent lexical
information \cite{Q32856000}.
A Wikidata-based embedding may benefit from such data.

\section{Acknowledgment}

The Danish Innovation Foundation supported this work through Danish
Center for Big Data Analytics driven Innovation (DABAI).

\begin{comment}
\url{https://github.com/idio/wiki2vec}

Profiling of web service.

As an example of the issues facing the matching, consider the word
``alcohol'' which, in a stringent 
interpretation, would mean the family of chemical compound, but in
common parlance would mean ethanol.
Some items are difficult to match to Wikidata items, e.g., Jackson,
accommodation and admission.

\end{comment}

% pdftotext Nielsen2017Wembedder.pdf ; wc Nielsen2017Wembedder.txt
% python -m scholia.tex write-bib-from-aux Nielsen2017Wembedder.aux

\bibliography{Nielsen2017Wembedder}
\bibliographystyle{acm}

\end{document}